\documentclass{article}

\usepackage{arxiv}

\usepackage[utf8]{inputenc} 
\usepackage[T1]{fontenc}    
\usepackage{hyperref}       
\usepackage{url}            
\usepackage{booktabs}       
\usepackage{amsfonts}       
\usepackage{nicefrac}       
\usepackage{microtype}      
\usepackage{graphicx}
\usepackage{lipsum}

\title{Unsupervised Sentiment Analysis of Plastic Surgery Social Media Posts}

\author{
  Alexandrea K. Ramnarine\\
  School of Professional Studies\\
  Northwestern University\\
  Chicago, IL 60611 \\
  \texttt{alexandrearamnarine2021@u.northwestern.edu} \\
}

\begin{document}
\maketitle

\begin{abstract}

The massive collection of user posts across social media platforms is primarily untapped for artificial intelligence (AI) use cases based on the sheer volume and velocity of textual data. Natural language processing (NLP) is a subfield of AI that leverages bodies of documents, known as corpora, to train computers in human-like language understanding. Using a word ranking method, term frequency-inverse document frequency (TF-IDF), to create features across documents, it is possible to perform unsupervised analytics, machine learning (ML) that can group the documents without a human manually labeling the data. For large datasets with thousands of features, t-distributed stochastic neighbor embedding (t-SNE), k-means clustering and Latent Dirichlet allocation (LDA) are employed to learn top words and generate topics for a Reddit and Twitter combined corpus. Using extremely simple deep learning models, this study demonstrates that the applied results of unsupervised analysis allow a computer to predict either negative, positive, or neutral user sentiment towards plastic surgery based on a tweet or subreddit post with almost 90\% accuracy. Furthermore, the model is capable of achieving higher accuracy on the unsupervised sentiment task than on a rudimentary supervised document classification task. Therefore, unsupervised learning may be considered a viable option in labeling social media documents for NLP tasks.
\end{abstract}

\keywords{Natural language processing \and unsupervised analysis \and social media \and Twitter \and Reddit \and plastic surgery}

\section{Introduction}
\label{sec:Intro}
\setlength{\parindent}{1em}
Cosmetic plastic surgery is an expensive yet increasingly popular set of procedures for both men and women, especially considering the ease of accessibility to surgeons, patient testimonials, and procedure-specific information, such as “before-and-after” visual aids, afforded by the Internet. The Internet is virtually a bottomless trove of textual data that are easily created at velocities of millions of posts per second across social media platforms. The exponential adoption of social media, largely facilitated by the wide global distribution of smart phone technology, is a known disseminator of beauty standards that is highly targeted to billions of users daily. Cosmetic surgery becomes a quick, permanent fix in adopting sought after beauty standards set by celebrities, models and social media “influencers”, relative to longer term alternatives such as dieting and exercise, or temporary alternatives such as adoption of fashion and cosmetic trends.

Social media, while distributing information about plastic surgery procedures, also provides a setting for public social commentary on the election of undergoing these surgeries. Users across many platforms are able to freely communicate their sentiments on a broad scale and even as granular as commenting on another individual’s posted surgical outcomes. Different social media platforms exist for different purposes, and thus attract and form distinct user bases that comprise these Internet communities. Each text post is unique to a user, time-stamped, geo- located, and has the capability to possess multiple data formats including but not limited to links and images. Therefore, text posts from social media sites provide specific insight into the public’s opinion on cosmetic surgery. It is thus reasonable to assume that the text posts made on one platform can be used to distinguish user-derived text from other platforms.

Curating massive corpora of text post documents from popular social media networks, Twitter and Reddit, is feasible with the implementation of AI web scraping technology. NLP is then leveraged to process and mathematically transform text to computationally understandable representations. ML methods can then identify patterns among the corpora that is an otherwise impossible task for a human to accomplish given the sheer volume of data. Deep learning (DL) methods, relying on powerful and speedy neural network technology, are then poised to use the NLP-curated and ML-processed data in order to accurately predict document class and user sentiment across the corpora. This study demonstrates that very simple, regularized neural network architectures effectively use unsupervised NLP to answer an easy to ask yet difficult to answer question, “how does the Internet feel about plastic surgery?”

\section{Literature Review}
\label{sec:LitRev}
\setlength{\parindent}{1em}
Opinion mining, better known as sentiment analysis, is an application of NLP that is particularly suited to extracting and assessing human opinion from textual data over the Internet and social media networks \cite{Singh2020}. While spoken language offers context surrounding feelings and opinions through auditory cues such as tone and pitch, written language often broadly captures polarity in discussions, which can be leveraged by AI. Trained AI are able to detect polarity, whether negative, positive, or neutral, based on word associations captured mathematically by distance metrics. Distance is able to represent and capture context, giving connotative rather than denotative meaning to the words that ultimately decide whether a word is positive or negative \cite{Miller_2015}. Therefore, ranking word importance to use as term features for AI is critical to achieve high accuracy for sentiment analysis, particularly unsupervised sentiment assignment. This study adopts the information retrieval ranking function of TF-IDF, combining two methods by \cite{Luhn1957} and \cite{Spärck2004} to assign weights to important terms across a corpus of documents.

Two popular unsupervised analyses are utilized in this study to support analyst judgment for assigning sentiment to social media posts that lack these labels. \cite{forgy65} and \cite{macqueen1967some} proposed the “k-means” method as an unsupervised algorithm to group observations based on minimizing variance, or the sum of squared deviations of points, to generate clusters of points using centroided means. \cite{blei} formulated LDA, which uses Bayesian probabilistic modeling to generate topic representations of NLP corpora based on their documents and associated terms. LDA therefore will ultimately support clustering analysis in generating labels for subsequent sentiment analysis.

More recently, \cite{Jelodar2020} applied LDA to extract features of YouTube comments, proposing that semantic topic extraction can directly aid in sentiment scoring of comments as “negative” or “positive” through an NLP-hybrid framework when applied to fuzzy lattice reasoning. In conjunction with unsupervised analysis, this application is useful in identifying user groups within social media networks. \cite{BennacerSeghouani2018} created an unsupervised approach to determine interests of social media users based on tweet semantics. A ML survey of unsupervised learning applications to NLP specifically highlights clustering algorithms such as k-means to address categorical outputs when labeled training data is not available for analytics.

\section{Methods}
\label{sec:Methods}
\setlength{\parindent}{1em}
\subsection{Data Acquisition}
A Python 3.8 development version of snscrape package was utilized to run command line bash scripts that scrape top and relevant social media posts from chosen Reddit subreddits and Twitter hashtags through March 2021, which serve as the document categories. Reddit queries from three different subreddits, “PlasticSurgery”, “CosmeticSurgery”, and “BotchedSurgeries”, had a maximum of 8000 or 4000 result scrapes of the latest posts based on total reported posts on each subreddit’s webpage. Twitter queries for each of the following hashtags, “plasticsurgery”, “liposuction”, “lipinjections”, “botox”, and “nosejob”, had a maximum of 8000 result scrapes of top tweets as determined by Twitter’s internal algorithm. Each scrape was saved as a JSON line file and subsequently read into respective Pandas dataframes. Null data were replaced with empty strings using NumPy. All Reddit and Twitter dataframes were concatenated respectively.

\subsection{Data Pre-processing}
The separate corpora dataframes were joined based on identification number, category, text, and title columns. Pre-processing steps for the combined corpus utilized a Python implementation of NLTK and custom functions to convert text to lowercase, remove punctuation, remove emojis and other Unicode characters, tokenize each document, remove English stop words, and stem each token. TF-IDF vectorization of the combined corpus employed additional pre-processing to drop duplicates and strip Unicode characters.

\subsection{Unsupervised Analysis}

The Scikit-learn TF-IDF Vectorizer was set to generate unigrams and subsequently fit to the combined corpus after randomly shuffling the dataframe. Scipy and Scikit-learn implementations of k-means using k of 8, 3, and 2, and a 2-component t-SNE multidimensionality rescale using cosine distance and perplexity of 50.0 were applied to the TF- IDF matrix. Each method underwent at least ten initializations and between 300 and 1000 iterations before convergence. The Scikit-learn implementation of LDA was used for topic modeling on the TF-IDF matrix, generating the top 20 words for 8, 3, and 2 topics. All visualizations were generated using MatPlotLib and Seaborn.

\subsection{Deep Learning}
The high-level Keras API on TensorFlow was utilized to build a Sequential dense neural network (DNN) with one input layer using rectified linear unit (ReLU) activation, one dropout regularization layer, and one output layer using softmax activation for both document category classification and sentiment analysis tasks. For sentiment analysis tasks, 1-D temporal convolutional (1D-CNN) Sequential models were built with an input layer of 32 units and a 3x3 kernel, ReLU activation and He uniform initialization, a 2x2 1-D max pooling layer, followed by a dropout and flatten layer feeding to a dense layer with 128 units before the final dense output layer. Each model was compiled using the Adam optimizer and a categorical cross-entropy loss function. After Scikit-learn 80\% train, 20\% test splitting of the TF-IDF matrix and labels, the models were fit to shuffled data and trained over 15 epochs with an internal training validation split of 20

For classification labels, each of the eight document categories were represented as an integer. For sentiment labels, analyst judgment of both the k-means clusters mapped into the t- SNE space and LDA topics was used to create three classes corresponding to negative, positive, or neutral sentiment. All labels were converted to categorical vectors using TensorFlow.

Training and validation loss and accuracy were tracked over each epoch before evaluating each model on the test sets. Scikit-learn classification reports of precision and recall, and confusion matrices were generated for each test instance.

\section{Results}
\label{sec:Results}

After vectorization of the combined corpus, unsupervised analyses were performed to visualize the distribution of document categories. Figure 1 illustrates the most similar documents towards the middle of the t-SNE space, while outlier documents sparsely separate at the fringe of the main cluster.

\begin{figure}[h]
\caption{t-SNE Dimensionality Reduction Mapped by Category}
\includegraphics[width=\textwidth]{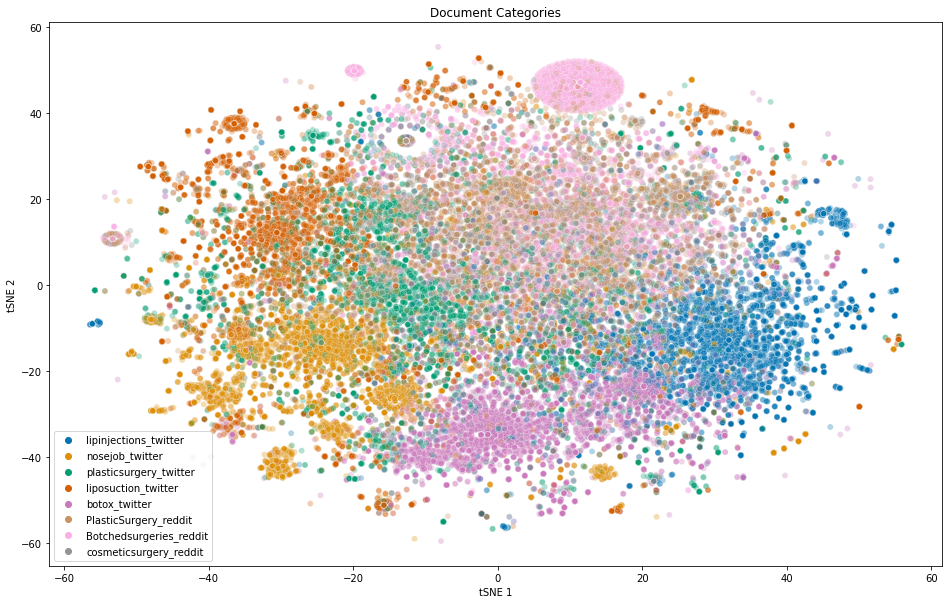}
\end{figure}

Documents from Reddit are more similar to each other than documents from Twitter, primarily falling in the mid to top right quadrant of the space while the Twitter documents cluster together along the bottom and left of the distribution. Documents related generally to plastic surgery, sourced from the plastic surgery Twitter hashtag or the Plastic Surgery and Cosmetic Surgery subreddits, are the most similar to each other and fall within the middle of the distribution. About half of the Botched Surgery subreddit documents strongly form a smaller cluster away from the rest, however the second half is well interspersed within the other subreddits toward the center of the distribution.

The liposuction and lip injection documents are most dissimilar from each other of the Twitter-sourced hashtags, while the nose job hashtag is similar to the liposuction and half of the Botox hashtag. The other half of the Botox hashtag is more similar to the lip injection hashtag source. This half of the Botox hashtag, along with the lip injection hashtag, are more dissimilar than the general plastic surgery hashtag than the nose job and liposuction hashtags.

The outlier documents are distributed in smaller yet distinguishable groups. There are three nose job Twitter hashtag outlier groups where two are somewhat related to the larger group, but one is more related to the main Botox hashtag group. There are many liposuction tweets that are more closely related to the Reddit documents than to Twitter documents. There is one Botched Surgery subreddit outlier group that is more related to the liposuction tweets. Finally, there are two strongly separated groups of mixed documents, however primarily comprised of Plastic Surgery and Botched Surgery subreddit documents. One of these falls within the main distribution but largely distanced in its entire circumference from the general plastic surgery tweets and subreddit posts, and the second falls completely out of the main distribution towards the most negative t-SNE 1 space, closer to a polarized outlier group of the twitter lip injection hashtag.

Centroid clustering was applied to the t-SNE space using a k-means approach. The eight generated clusters highlight the strongest outlier document groups, the differences among the twitter hashtags, and the similarities among both the Reddit documents and the general plastic surgery documents, as illustrated in Figure 2.

\begin{figure}[h]
\caption{k-means Clustering Mapped to t-SNE Space}
\includegraphics[width=\textwidth]{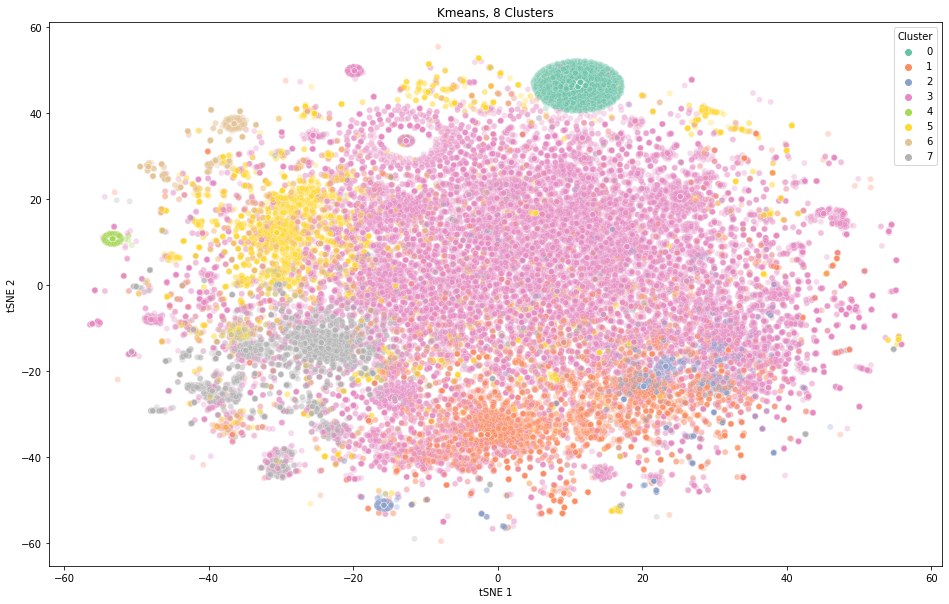}
\end{figure}

Cluster 3 is a “catch all”, predominantly mapping to the Reddit documents and general plastic surgery related documents. Cluster 0 mapped directly to the Botched Surgery outlier group, cluster 1 to the Botox hashtag, cluster 7 to the nose job hashtag, and cluster 5 to the liposuction hashtag. Cluster 2 seemed to map to documents that fell directly central within the Botox and lip injection document space but were not sourced from either of those hashtags. Clusters 3 and 6 highlight strong outlier groups in the t-SNE space, where the former maps to the outliers of the general plastic surgery documents, and the latter maps to the fringe outliers of the liposuction hashtag tweets.

In order to stratify the documents into three balanced groups corresponding to positive, negative, or neutral sentiment, k-means clustering was performed again on the TF-IDF matrix using k equal to 3 and 2. Appendix A demonstrates that the most significant differences between the documents using centroid clustering is between the main distribution and the large Botched Surgery outlier group. Therefore, LDA was employed to further support analyst judgement in unsupervised sentiment labeling. Figure 3 depicts the results of the top 20 terms for eight topics across the combined corpus, corresponding to the eight document categories.

\begin{figure}[h]
\caption{Top 20 Terms by LDA Topic Modeling}
\includegraphics[width=\textwidth]{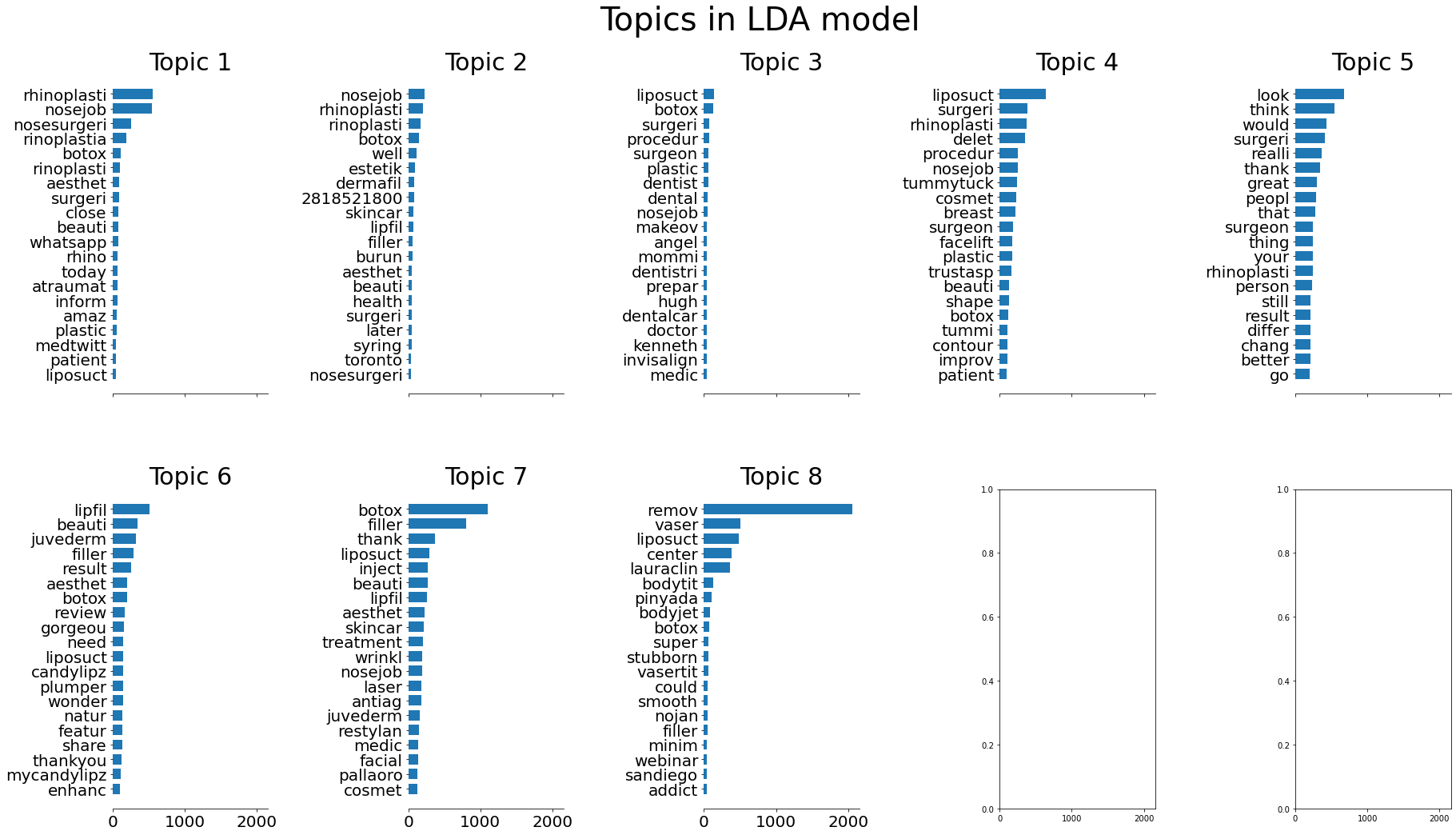}
\end{figure}

\begin{table}[h!]
 \caption{Selected k-means Top Terms}
  \centering
\begin{tabular}{llllllll}
Cluster 0                                                                                                                     & Cluster 1                                                                                                             & Cluster 2                                                                                                        & Cluster 3                                                                                                           & Cluster 4                                                                                                    & Cluster 5                                                                                                        & Cluster 6                                                                                                                       & Cluster 7                                                                                                                 \\ \hline
\begin{tabular}[c]{@{}l@{}}remove\\ botch\\ excess skin\\ evil\\ evildick\\ evilqueen\\ exaggerate\\ exboyfriend\end{tabular} & \begin{tabular}[c]{@{}l@{}}inject\\ skincare\\ filler\\ treatment\\ facial\\ antiaging\\ botox\\ wrinkle\end{tabular} & \begin{tabular}[c]{@{}l@{}}fabulous\\ gratitude\\ motivate\\ fit\\ well\\ comfort\\ lip goal\\ love\end{tabular} & \begin{tabular}[c]{@{}l@{}}thank\\ gorgeous\\ happiness\\ amaze\\ wonder\\ please\\ pretty\\ candylipz\end{tabular} & \begin{tabular}[c]{@{}l@{}}delete\\ ban\\ swollen\\ swell\\ evil\\ excess\\ exboyfriend\\ didnt\end{tabular} & \begin{tabular}[c]{@{}l@{}}remove\\ attack\\ stubborn\\ improve\\ problem\\ eliminate\\ double-chin\end{tabular} & \begin{tabular}[c]{@{}l@{}}(contains \\ numerical \\ terms, like \\ injection \\ doses \\ and \\ phone \\ numbers)\end{tabular} & \begin{tabular}[c]{@{}l@{}}surgeon\\ surgery\\ beauty\\ medic\\ cosmetic\\ rhinoplasty\\ procedure\\ patient\end{tabular}
\end{tabular}
\label{tab:table}
\end{table}

While the majority of the top terms can be considered neutral, there are a few that can be mapped back to the k-means top term results, shown in Table 1, in order to assign sentiment labels to each k-means cluster. From Topic 4, “delete” and “improve”, and from Topic 8, “remove”, “addict”, and “stubborn”, are key terms indicative of negative sentiment if also highlighted by k-means. Reducing the LDA to three or two topics still captures these negative connotation terms. Therefore, k-means clusters 0, 4, and 5 were assigned a negative sentiment label, clusters 1, 6, and 7 were assigned a neutral sentiment label, and clusters 2 and 3 were assigned a positive sentiment label. To correct for class imbalance, any Botched Surgery subreddit documents not assigned to negative sentiment were reassigned a negative label based on that particular subreddit’s culture of mocking and shaming plastic surgery procedure outcomes subjectively deemed poor.

\subsection{Predicting on Supervised versus Unsupervised Labels}
A very simple one-layer DNN architecture, utilizing 30\% dropout regularization, was used to test supervised document category classification versus the unsupervised sentiment analysis. Appendix C illustrates that training accuracy increases with epochs; however,
validation accuracy stagnates. Training loss decreases with epochs, but validation loss increases in both classification and sentiment analysis cases. Table 2 compares the performances between classification and sentiment analysis tasks on the combined corpus.

\begin{table}[h!]
 \caption{Dense Neural Network Performance}
  \centering
  \begin{tabular}{lllllll}
    \toprule
    \multicolumn{1}{c}{} & \multicolumn{2}{c}{Training}                            & \multicolumn{2}{c}{Validation}                          & \multicolumn{2}{c}{Test}                                \\ \cline{2-7} 
Task                 & \multicolumn{1}{c}{Accuracy} & \multicolumn{1}{c}{Loss} & \multicolumn{1}{c}{Accuracy} & \multicolumn{1}{c}{Loss} & \multicolumn{1}{c}{Accuracy} & \multicolumn{1}{c}{Loss} \\ \hline
Classification       & 93.13                        & 0.1767                   & 78.16                        & 0.7555                   & 77.78                        & 0.7859                   \\
Sentiment            & 97.77                        & 0.0473                   & 87.84                        & 0.4652                   & 87.12                        & 0.4768\\
    \bottomrule
    \emph{Note:} Accuracy shown in percent.
  \end{tabular}
  \label{tab:table}
\end{table}

Overall, the model was able to achieve better performance on unsupervised sentiment analysis versus supervised document classification. Appendices D and E compare the harmonic mean and confusion matrices of the two learning tasks. For classification, there was a class imbalance for the lip injections Twitter hashtag, therefore the F1-score was low and misclassification rate was high relative to the performance on the other labels. The model performed best on correctly classifying the nose job Twitter documents.

For sentiment analysis, although there was class imbalance skewed towards over-representation of the positive-labeled documents, this had no noticeable effect on model performance. The model performed best on predicting neutral sentiment and relatively worse on predicting negative sentiment, however the precision and recall metrics between the three sentiments are similar. Almost all of the neutral documents were correctly predicted as such, and less than 20

\subsection{Sentiment Analysis}
Given that a simple DNN could achieve near 90\% accuracy on unsupervised sentiment analysis, experiments varying dropout regularization and use of a temporal convolutional neural network were conducted. Table 3 summarizes the training, validation, and test results of these experiments to predict sentiment.

\begin{table}[h!]
 \caption{Model Regularization and Architecture Experiments for Sentiment Prediction}
  \centering
\resizebox{\columnwidth}{!}{%
\begin{tabular}{lccccccc}
    \toprule
 & \multicolumn{1}{l}{} & \multicolumn{2}{c}{Training} & \multicolumn{2}{c}{Validation} & \multicolumn{2}{c}{Test} \\ \cline{3-8} 
Model  & Dropout & Accuracy & Loss   & Accuracy & Loss   & Accuracy & Loss   \\ \hline
DNN    & 0       & 97.33    & 0.0567 & 86.98    & 0.5057 & 86.27    & 0.5227 \\
DNN    & 0.3     & 97.77    & 0.0473 & 87.84    & 0.4652 & 87.12    & 0.4768 \\
DNN    & 0.6     & 96.41    & 0.0830 & 88.26    & 0.3627 & 87.53    & 0.3729 \\
1D-CNN & 0       & 98.64    & 0.028  & 87.59    & 0.4912 & 87.08    & 0.5292 \\
1D-CNN & 0.3     & 98.61    & 0.0311 & 87.12    & 0.5203 & 86.92    & 0.5604 \\
1D-CNN & 0.6     & 98.40    & 0.0388 & 88.80    & 0.5048 & 88.21    & 0.5552\\
    \bottomrule
    \emph{Note:} Dropout rate shown, accuracy shown in percent.
  \end{tabular}%
  }
\end{table}

Increasing dropout rate in both model cases increases both validation and test accuracies overall. However, increasing dropout rate for the 1D-CNN does not improve validation or test loss compared to using no dropout regularization, and instead caused the validation loss to behave erratically over training epochs (Appendix G). Using dropout had no substantial effect on validation accuracy over epochs of the 1D-CNN. Appendix F illustrates that using high dropout rates for the DNN shrinks the gap between training and validation metrics at each epoch, notably shrinking validation loss despite the similar upwards trending loss over epochs in both zero and 60\% dropout cases.

Appendix H displays the test results of the sentiment analysis comparing dropout regularization between the two models. The 1D-CNN using 60\% had the highest true classification rate of positive sentiment, while the 1D-CNN using 30\% dropout had the lowest. The DNN using 60\% dropout had the best classification rate of negative sentiment, while the DNN using no dropout performed relatively poorly on correctly classifying negative sentiment. The 1D-CNN using 30\% dropout correctly predicted neutral sentiment at the highest rate, and the DNN using no dropout correctly predicted neutral sentiment at the lowest relative rate among the models. All models displayed almost negligible misclassification rates bidirectionally between negative and neutral sentiment.

\clearpage
\newpage
\section{Discussion}
\label{sec:Discussion}
\setlength{\parindent}{1em}
While it is reasonable to assume that virtual communities formed over social media forums tend to attract like-minded opinions, this over-generalization may conflate the outlier user posts within each group. Herein, it is demonstrated that applying unsupervised dimensional reductions and clustering algorithms to an extremely large and heterogenous corpus of Twitter and Reddit text data is a viable option to capture user sentiment based on word rankings. In conjunction with topic modeling, these methods may be employed to label noisy text data in a semi-supervised manner.

The Twitter and Reddit documents mostly separate in the t-SNE space, suggesting that the types of posts, and therefore the user bases, can distinguish between the two social media networks. The relative homogeneity of the Reddit to Twitter distribution supports the idea that Reddit posts, and therefore users, are somewhat more similar to each other that Twitter users. This is likely a function of subreddits being niche internet communities with users sharing multiple posts within the same subreddit, and perhaps even between the three sampled plastic surgery subreddits since there are no other major related communities that were found on Reddit pertaining to plastic or cosmetic surgery. It is a fair assumption that Twitter has a more heterogenous representation because hashtags do not act as niche communities the way subreddits are structured to.

Given that the sourced tweets are stratified mainly by procedure related hashtags, it is unsurprising that non-invasive, injection-based procedures cluster closely together, such as the lip injection and Botox clusters, while the invasive procedures such as nose job and liposuction cluster together. k-means cluster 1 therefore must be representative of non-invasive or injection- based procedures. That nose job and Botox documents are still relatively close in distance in the t-SNE space indicates relatedness due to terms associated with facial procedures. Interestingly, these four hashtags for multiple smaller outlier clusters in the t-SNE space, probably indicative of underlying sentiment distributions given the k-means mapped analysis and strong predictive power of the neutral networks. Despite the biased sourcing used for these tweets, the general plastic surgery Twitter hashtag documents almost uniformly span all of the procedure-specific Twitter documents in the t-SNE space.

Surprisingly, the unsupervised generated labels seemingly allowed the simplest of neural networks to outperform a supervised NLP task, which may suggest that the content of plastic surgery related documents sourced from Twitter and Reddit are better captured by analyst judgment sentiment and not by the empirical document source. This further suggests that the term ranking methodology employed, together with topic modeling, generated strong indicators of social media user opinion, effectively grouping words based on cosine distance.

Using a temporal convolutional network, a model theoretically better suited to capturing high and low dimensionalities of sequential text data, showed negligible improvement over the DNN in terms of accuracy, loss, and sentiment true classification rate. In general, both neural networks overfit the training data, averaging about 10\% differences in accuracy between training and test instances. Training and validation instances indicate that both models would benefit from early stopping well before 10 epochs in order to achieve higher validation accuracy and lower validation loss; it is assumed that the test metrics would follow in suit.

All models had comparably high precision and recall for both neutral and positive sentiment, although the F1-scores for negative sentiment prediction were not dramatically lower. Given absolute true classifications, the models overall were able to distinguish negative from neutral sentiment very well. For each model, most of the misclassifications occurred between positive and negative sentiment, followed by positive and neutral sentiment. The top-ranking terms therefore must strongly segregate neutral from other sentiment in the case of plastic surgery, indicative of volume of terms used and associated with the medical procedures rather than with user opinions of those procedures or their outcomes. That the models struggled most with misclassifications between positive and negative sentiment could be indicative of vernacular and colloquial usage of terms mixed with denotative usage, confounding learning and thus impairing the decision boundary between these two sentiments. Additionally, it may be useful to use n-grams rather than unigrams to better define terms, such as “beauty” and “change”, that could realistically fall into any sentiment category for plastic surgery depending on the context it is used in.

The high predictive capacities of these simple models indicate that favored NLP recurrent neural networks (RNNs), including gated recurrent unit networks and long short-term memory networks, may not perform that much better for sentiment analysis of these social media sourced documents, given the abbreviated length of each document and the frequently associated vanishing gradient problems with RNNs. While it may be interesting to pursue future work with different model architectures, the results from the temporal convolutional network, considered an advancement to simple RNNs given its ability to capture spatio-temporal information, indicate that it may be better to invest efforts in curation, vectorization and thus representation of top terms, using fewer terms but a more polarizing vocabulary to model the data after. Additionally, it may be fruitful to capture a wider breadth of hashtags from Twitter, more posts from the subreddits, and even venture to other social media networks for relevant plastic surgery documents to expand the user base, and thus opinion, representation.

\bibliographystyle{unsrt}  
\bibliography{references}  






\end{document}


\maketitle

\clearpage
\newpage
\section{Appendix A}
\label{sec:A}
\emph{k-means Clustering Mapped to t-SNE Space for k=3 \& k=2}
\begin{figure}[ht!]
\includegraphics[width=\textwidth]{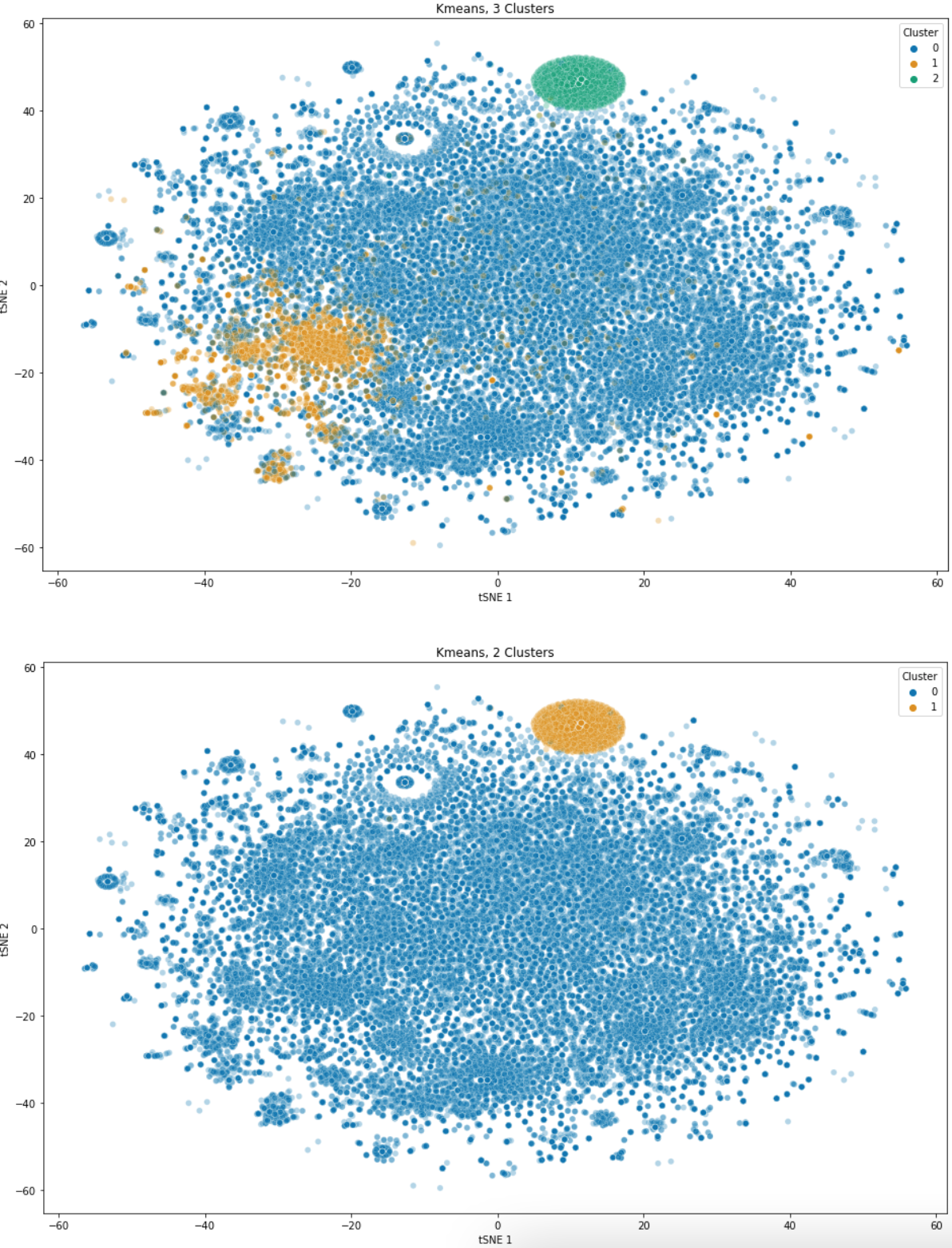}
\end{figure}

\clearpage
\newpage
\section{Appendix B}
\label{sec:B}
\emph{Top 20 LDA Terms for 3 \& 2 Topics}
\begin{figure}[ht!]
\includegraphics[width=\textwidth]{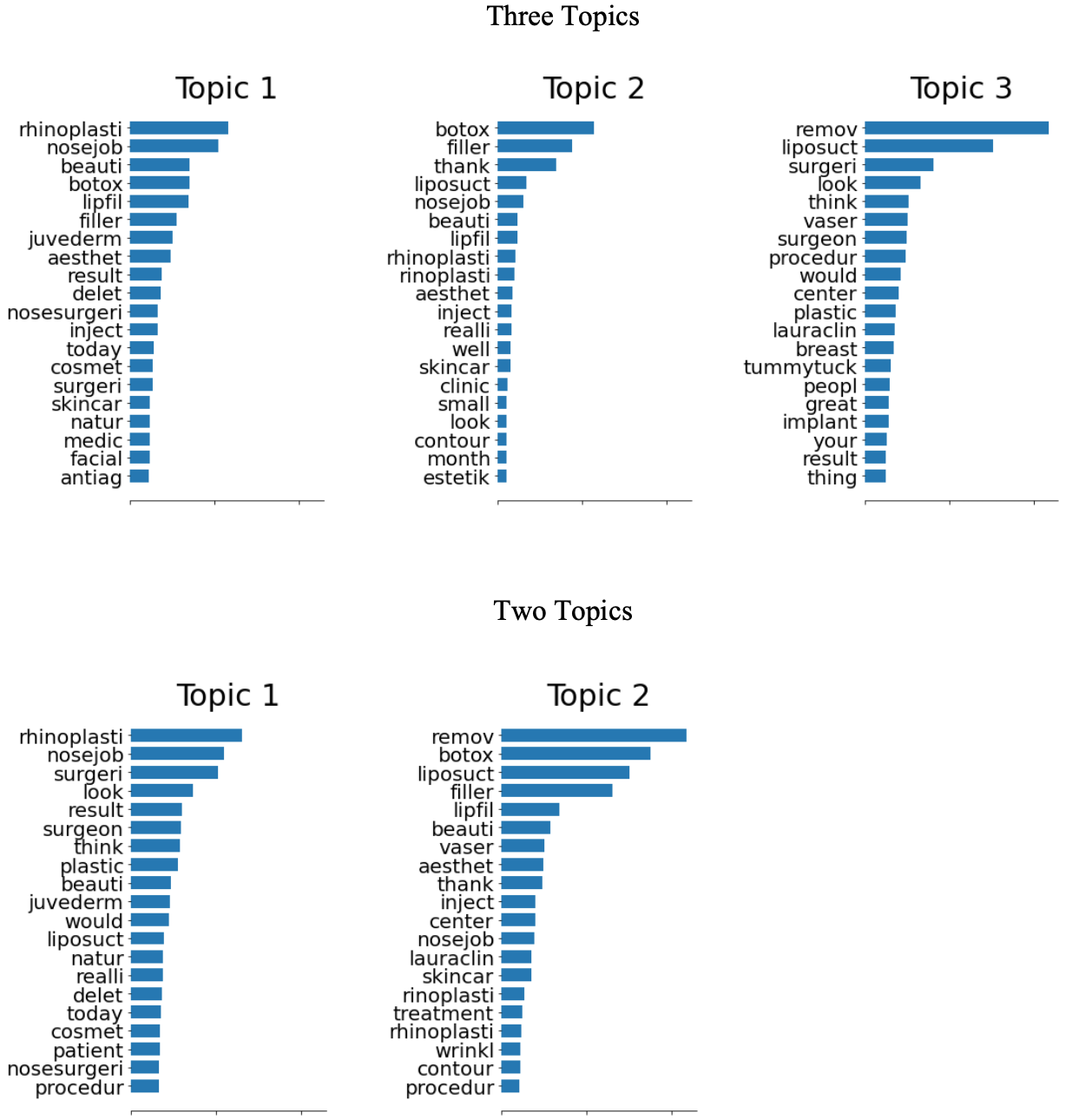}
\end{figure}

\clearpage
\newpage
\section{Appendix C}
\label{sec:B}
\emph{Dense Neural Network Training \& Validation Metrics}
\begin{figure}[ht!]
\includegraphics[width=\textwidth]{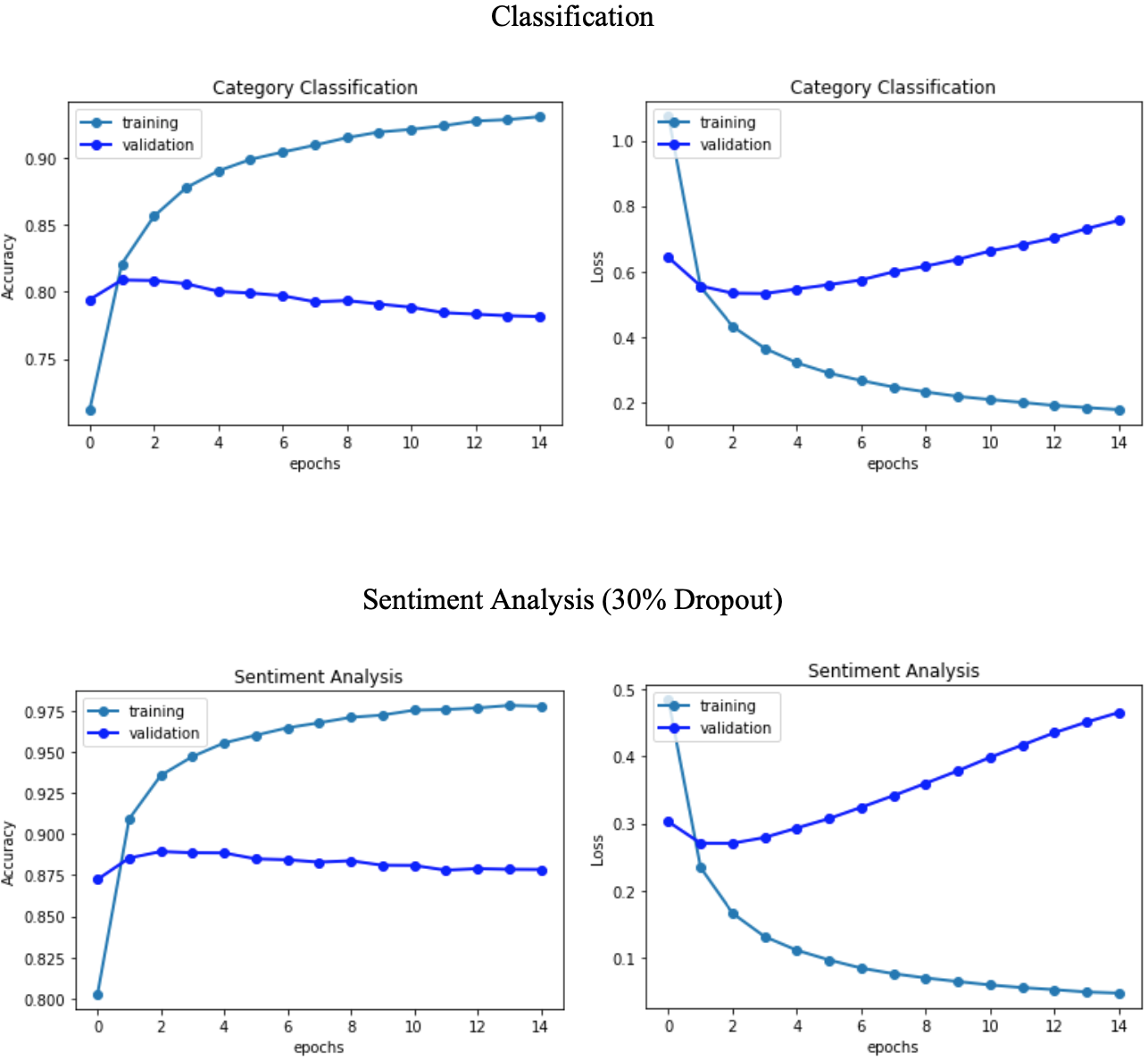}
\end{figure}

\clearpage
\newpage
\section{Appendix D}
\label{sec:B}
\emph{Classification Test Evaluation}
\begin{figure}[ht!]
\includegraphics[width=\textwidth]{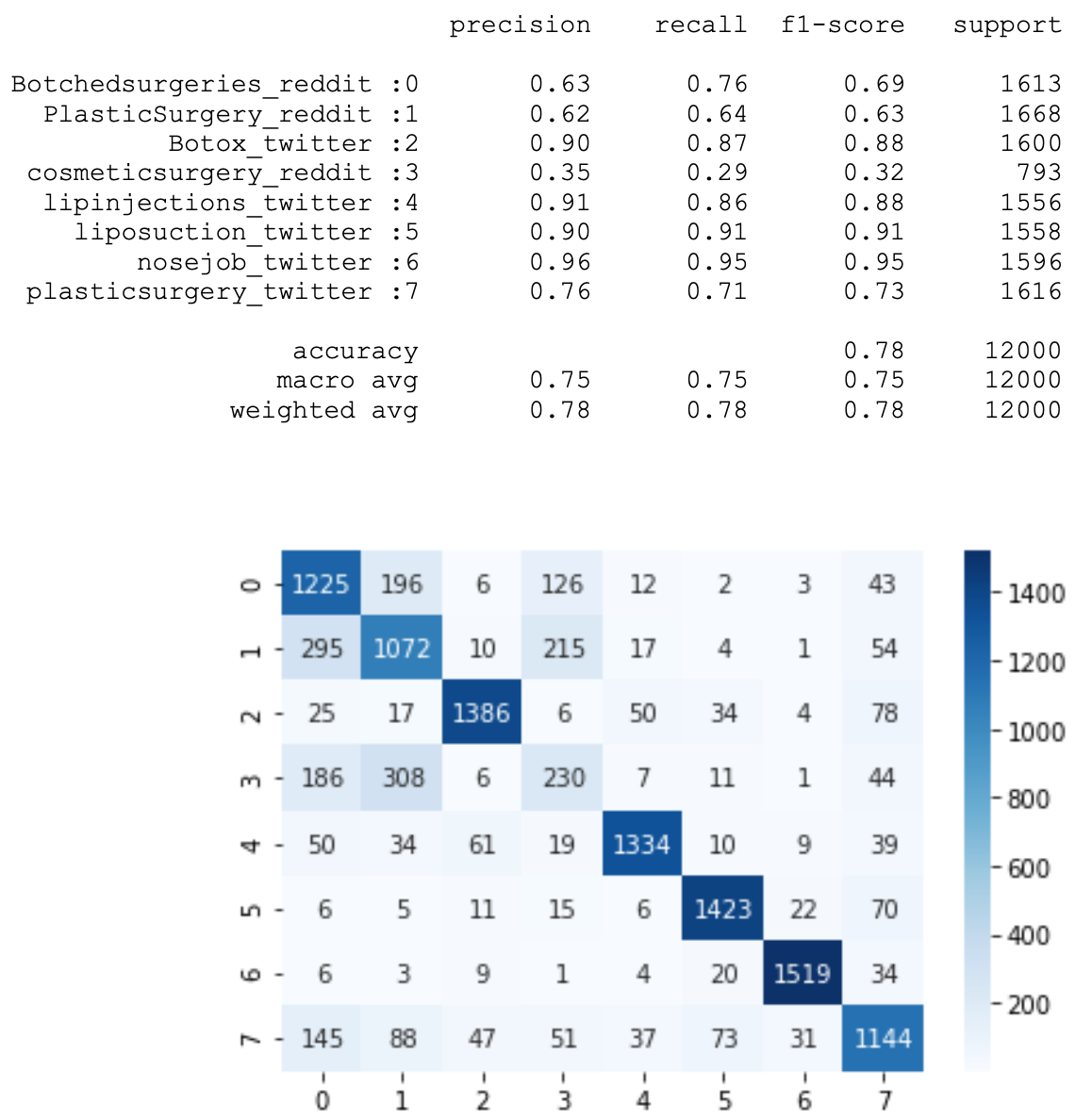}
\end{figure}

\clearpage
\newpage
\section{Appendix E}
\label{sec:B}
\emph{Sentiment Analysis (30\% Dropout) Test Evaluation}
\begin{figure}[ht!]
\includegraphics[width=\textwidth]{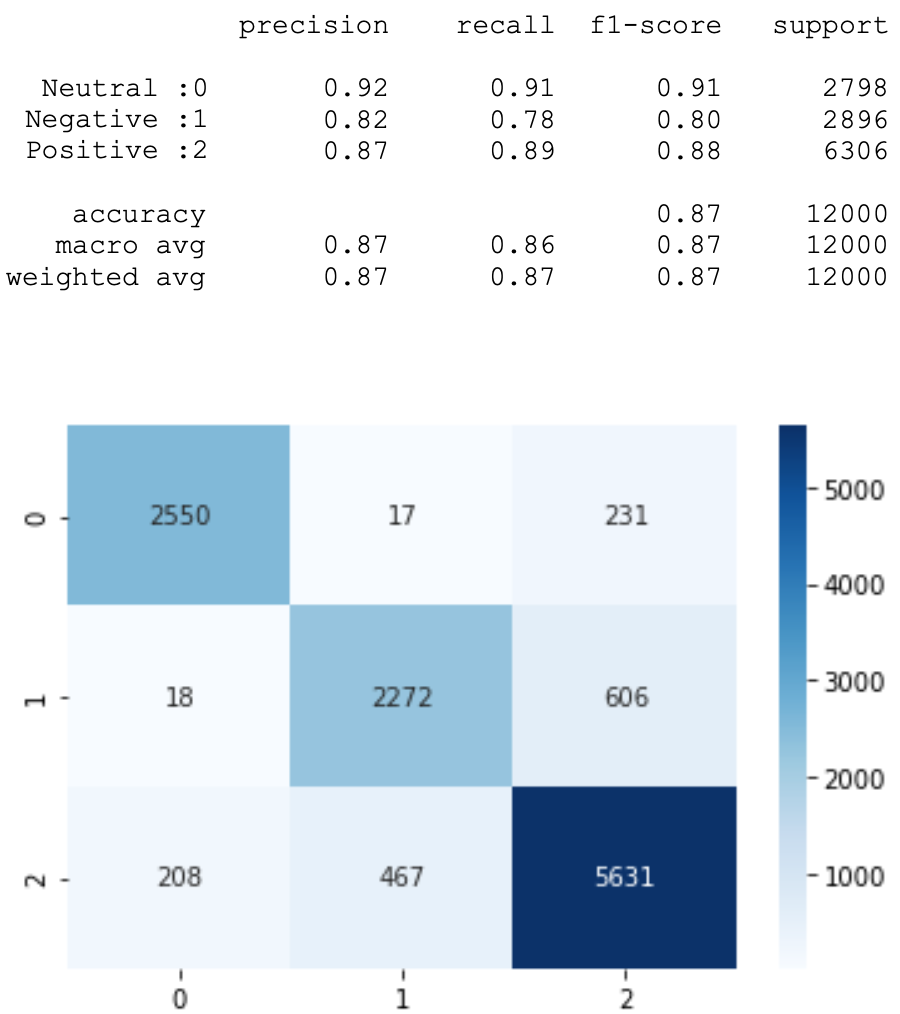}
\end{figure}

\clearpage
\newpage
\section{Appendix F}
\label{sec:B}
\emph{DNN Dropout Regularization Comparison of Training \& Validation Metrics}
\begin{figure}[ht!]
\includegraphics[width=\textwidth]{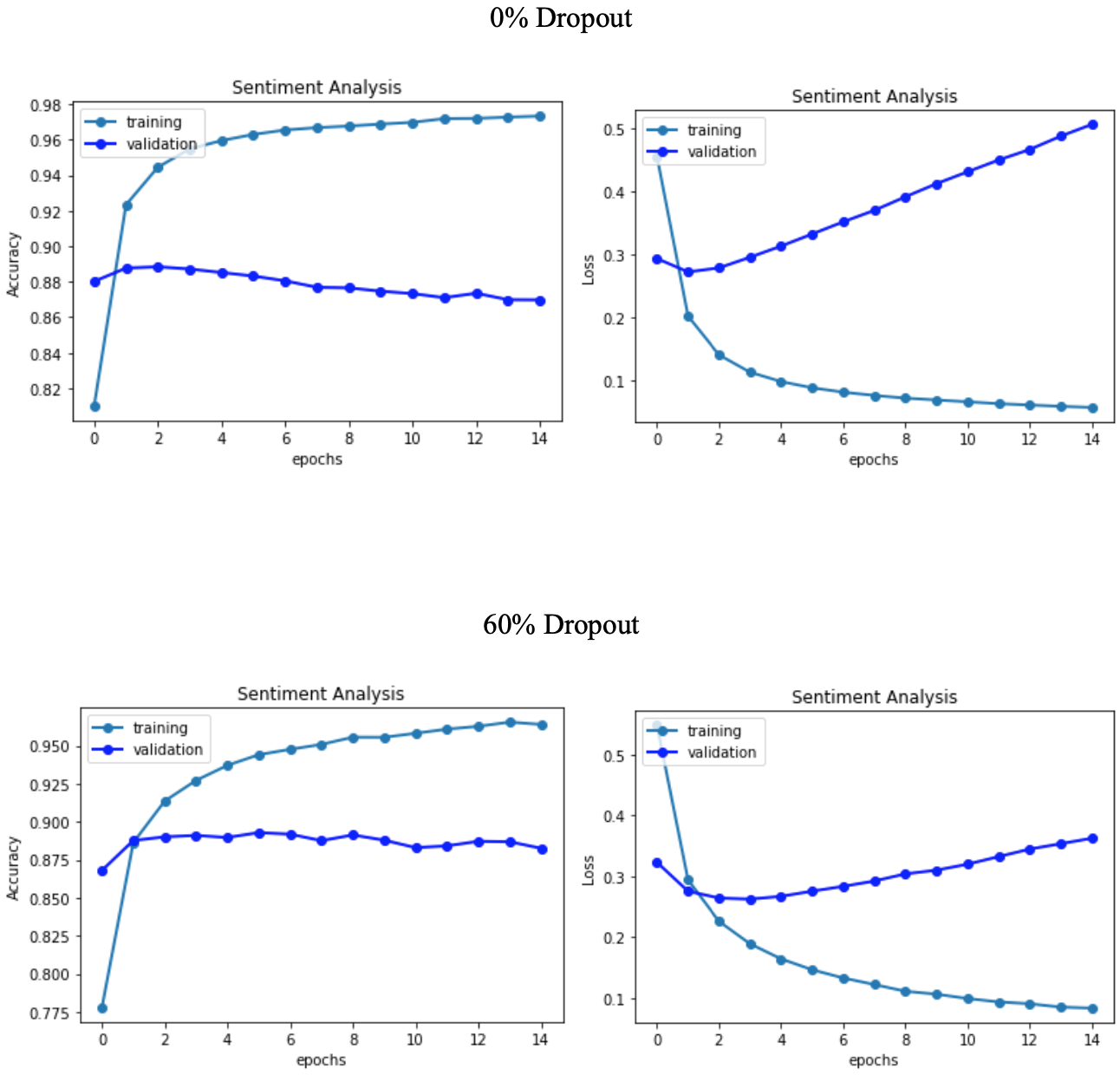}
\end{figure}

\clearpage
\newpage
\section{Appendix G}
\label{sec:B}
\emph{1D-CNN Dropout Regularization Comparison of Training \& Validation Metrics}
\begin{figure}[ht!]
\includegraphics[width=\textwidth]{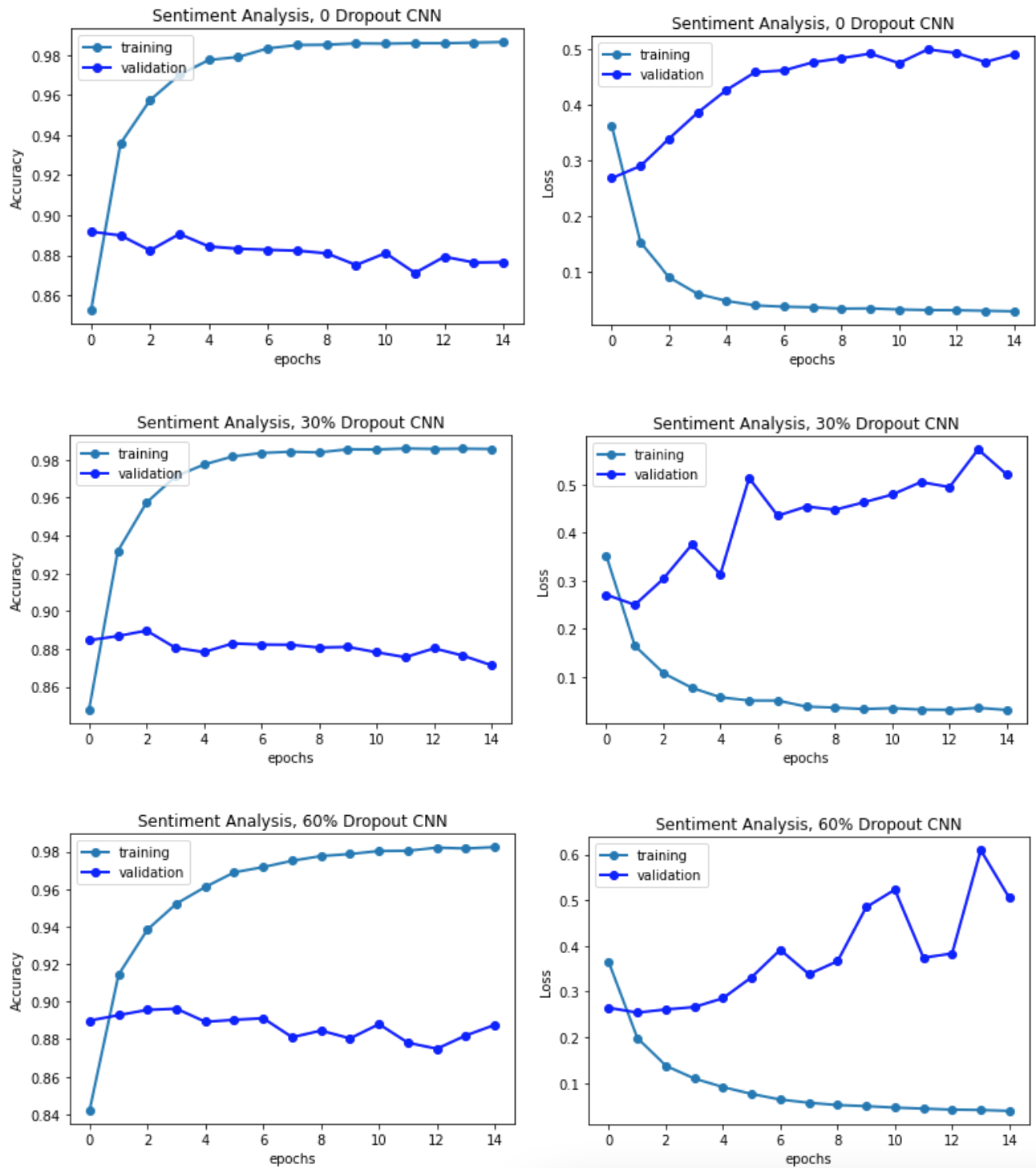}
\end{figure}

\clearpage
\newpage
\section{Appendix H}
\label{sec:B}
\emph{Sentiment Analysis Test Results Comparing Model Architecture and Dropout Regularization}
\begin{figure}[ht!]
\includegraphics[width=\textwidth]{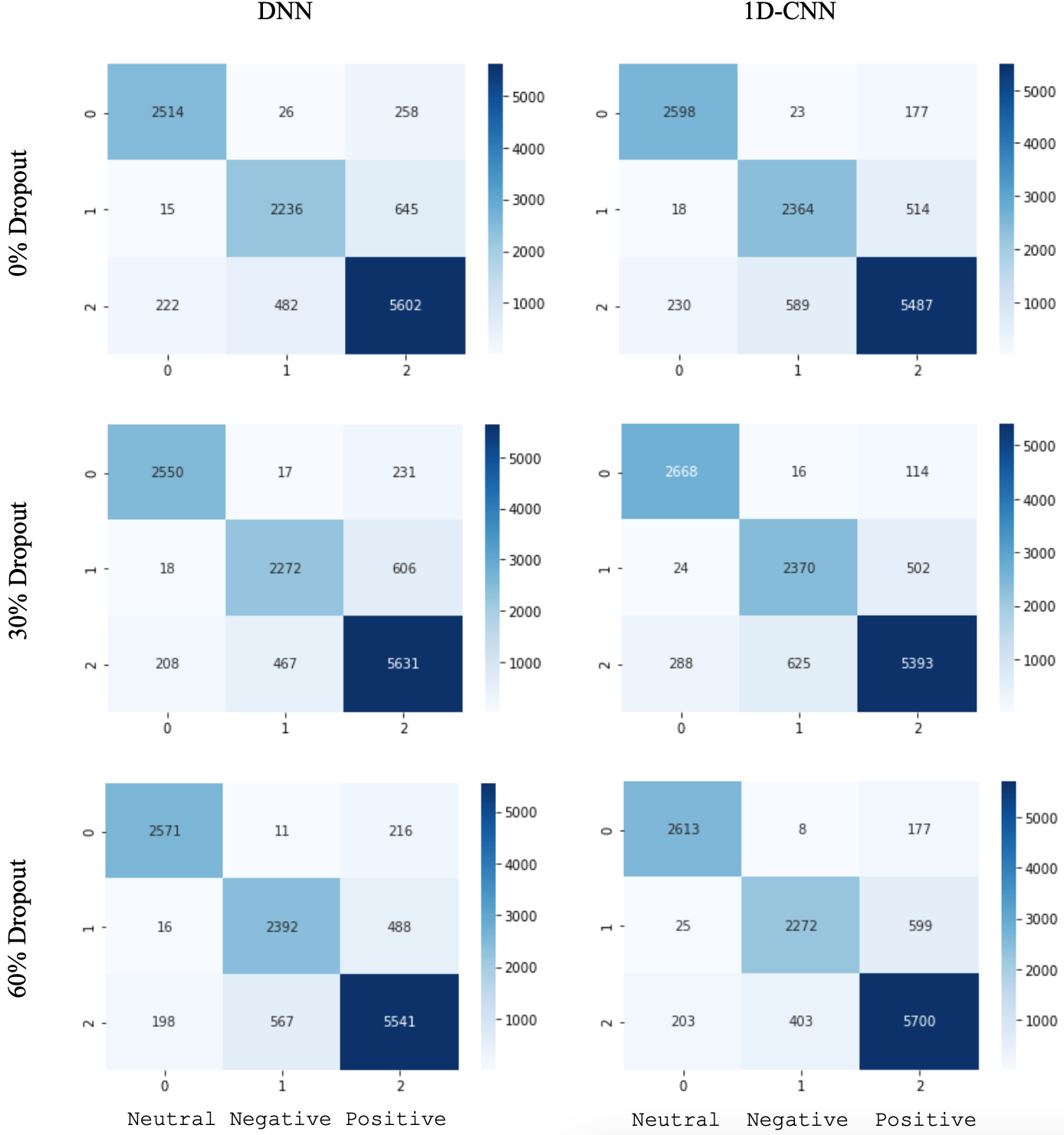}
\end{figure}